\documentclass[lettersize,journal]{IEEEtran}
\usepackage[utf8]{inputenc}
\usepackage[T1]{fontenc}
\usepackage{amsmath}
\usepackage{amsfonts}
\usepackage{amssymb}
\usepackage{graphicx}
\usepackage{enumitem}
\usepackage{comment}
\usepackage{wrapfig}
\usepackage{soul}
\usepackage[dvipsnames]{xcolor}
\usepackage{xpatch}
\usepackage{cite}
\usepackage{multirow}
\usepackage{tabularx}
\usepackage[justification=centering]{caption}
\usepackage[resetlabels,labeled]{multibib}
\usepackage{subcaption}
\usepackage[ruled,norelsize]{algorithm2e}

\newcites{O}{Produced Research Output}

\newcommand*{\newsub}[1]{\textcolor{black}{#1}}

\newcommand{\set}[1]{\mathbf{#1}}

\makeatletter
\def\endthebibliography{%
	\def\@noitemerr{\@latex@warning{Empty `thebibliography' environment}}%
	\endlist
}
\makeatother

\begin{document}
	\title{Federated Hyperdimensional Computing for Resource-Constrained Industrial IoT}
    \author{%
    Nikita~Zeulin, Olga~Galinina, Nageen~Himayat, and~Sergey~Andreev%
    \IEEEcompsocitemizethanks{\IEEEcompsocthanksitem N. Zeulin and S. Andreev are with  Tampere University, Tampere, Finland.\protect\\
    E-mail: \{nikita.zeulin,sergey.andreev\}@tuni.fi
    \IEEEcompsocthanksitem O. Galinina is with Tampere Institute for Advanced Study, Tampere University, Finland. E-mail: olga.galinina@tuni.fi.\protect\noindent
    \IEEEcompsocthanksitem N. Himayat is with Intel Corporation, Santa Clara, CA, United States. E-mail: nageen.himayat@intel.com}\protect\\
    }
	\pagestyle{empty}
	\maketitle
	\thispagestyle{empty}
	
	\begin{abstract}
	
 \newsub{In the Industrial Internet of Things (IIoT) systems, edge devices often operate under strict constraints in memory, compute capability, and wireless bandwidth. These limitations challenge the deployment of advanced data analytics tasks, such as predictive and prescriptive maintenance.  
 In this work, we explore hyperdimensional computing (HDC) as a lightweight learning paradigm for resource-constrained IIoT. Conventional centralized HDC leverages the properties of high-dimensional vector spaces to enable energy-efficient training and inference. We integrate this paradigm into a federated learning (FL) framework where devices exchange only prototype representations, which significantly reduces communication overhead. 
 Our numerical results highlight the potential of federated HDC to support collaborative learning in IIoT with fast convergence speed and communication efficiency. These results indicate that HDC represents a lightweight and resilient framework for distributed intelligence in large-scale and resource-constrained IIoT environments.}

\end{abstract}
	
\section{Introduction}

Industrial Internet of Things (IIoT) and machine learning (ML) form the backbone of modern data-driven industrial solutions. These technologies enable predictive and prescriptive maintenance, which uses data to optimize operations and prevent equipment failures. 
An IIoT system may involve thousands of small energy-constrained devices equipped with multiple sensors and connected through a wireless network. Sensor readings are subsequently passed into an ML algorithm for training, fine-tuning, or inference. Depending on the application, these tasks can be performed either on the edge server or directly on the IIoT device. For example, mobile robots may use GPU-based hardware platforms for on-device object recognition, while smart sensors offload their measurements to an edge server for further processing.

In large-scale IIoT scenarios, traditional centralized and on-device ML approaches face inherent limitations. Centralized systems may struggle with the high communication overhead to transmit large volumes of data from multiple devices. 
In turn, on-device ML methods are constrained not only by the limited computational resources, memory, and device battery but also by potentially biased data distributions, which can lead to poorly generalizable models. 

A popular solution to address these challenges is to apply \emph{federated learning} (FL) techniques. 
FL is a distributed learning approach that allows multiple devices to collaboratively train a common ML model by exchanging their local updates without disclosing the raw data. Such collaboration can occur within a single industrial facility, across multiple sites of the same company, or even among organizations operating similar equipment. Importantly, applying FL in these cases helps prevent the explicit disclosure of the potentially sensitive data of the manufacturer.

{IIoT devices remain inherently resource-constrained, with limited processing and storage capabilities and energy budgets. Moreover, tasks like predictive and prescriptive maintenance are often secondary 
to core industrial operations and therefore should operate with minimal bandwidth and power overhead. As a result, federated IIoT solutions should be designed to operate efficiently within these constraints.}

The community has explored various strategies for improving the efficiency of FL system. 
The authors of the survey in \cite{imteaj2021survey} emphasize that energy efficiency and low memory consumption are essential for on-device FL training, although resource-intensive neural network (NN) models are the most popular among the reviewed works. Communication constraints can be alleviated via model sparsification, 
quantization, 
or subsampling FL participants.
Most of these methods, however, are tailored for NN-based models, the computational complexity of which can be prohibitively high for resource-constrained IIoT devices.

Lightweight models such as support vector machines (SVMs) and gradient tree boosting algorithms like XGBoost 
are often used in IIoT systems but may also face limitations in distributed settings. For example, distributed training of XGBoost has very high communication overhead in horizontal FL scenarios, where participants share a common feature space but have distinct data samples. Federated training requires communicating the gradient trees with the total number of features scaling as $O(2^L)$, where the tree depths $L$ is between $8$ and $500$ in practice \cite{ma2023gradient}.
Training feature-mapped SVM (a replacement of kernelized non-linear SVMs in federated settings) requires solving distributed optimization problems in high-dimensional spaces, which can be computationally demanding. Despite being lightweight, these FL solutions may still remain challenging to deploy on small and resource-constrained IIoT devices in large-scale wireless networks.

\emph{Hyperdimensional computing} (HDC), a computational paradigm based on representing information using high-dimensional vectors and simple vector operations, offers a promising alternative for energy-efficient ML \cite{kanerva2009hyperdimensional,imani2021revisiting}. Its hardware-oriented architecture enables cost-effective training and inference with competitive performance to NN models on time series data \cite{schlegel2025structured}, graphs \cite{kang2025relhdx}, and images \cite{yun2024neurohash}. This efficient design of HDC algorithms makes them particularly attractive for IIoT applications that require fast learning with minimal resource consumption.

Most existing FL implementations rely on gradient-based training, such as FedAvg and its extensions. In contrast, this work explores a fundamentally different learning paradigm that avoids gradient exchange. HDC replaces model-weight synchronization with prototype aggregation, which fundamentally changes the computational and communication structure of the federated system. Some recent FL methods have also explored prototype-based model exchange, typically based on NN feature extractors. In contrast, HDC directly learns class prototypes in a high-dimensional symbolic space using simple vector operations.

This work illustrates that collaborative HDC learning can be realized across large populations of resource-constrained IIoT devices without relying on gradient-heavy models or high communication overhead. The remainder is organized as follows. In the next section, we introduce the key principle of HDC. 
In Section III, we outline the core challenges of IIoT systems that employing HDC can address. In Section IV, we
present a federated HDC framework and provide a numerical evaluation of its accuracy and reduction of resource utilization in Section V. We conclude by discussing key insights and open research problems in Section VI.

\begin{figure}
    \centering
    \includegraphics[width=0.95\linewidth]{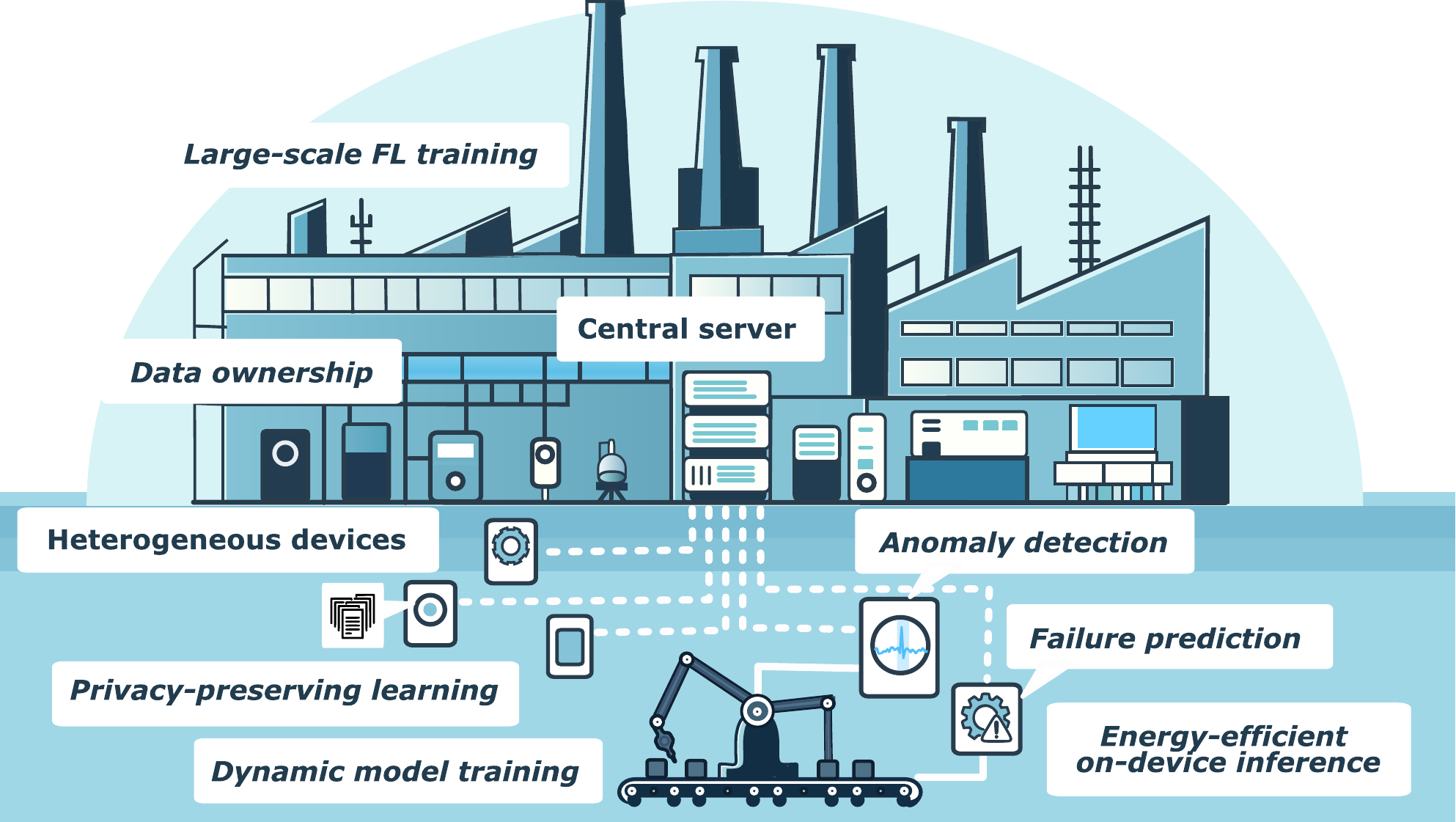}
    \caption{Illustration of envisioned industrial IoT system.}
\end{figure}

\section{Principles of Hyperdimensional Computing}
\label{sec:hdc_principles}

\newsub{Hyperdimensional computing (HDC) is a relatively new concept in FL and has the potential to address the resource constraints and computational challenges in IIoT environments. To provide context, this section outlines the core principles of HDC data representation and key procedures that make HDC suitable for resource-constrained IIoT.
}

\subsection{Foundational Concepts}
HDC is a pattern recognition concept inspired by the principles of neural activity of biological organisms. The key idea of HDC is to map, or \textit{encode}, the input data into randomized vectors of very large dimensionality (e.g., 5-10K features), known as hyperdimensional (HD) vectors, or \textit{hypervectors} \cite{kanerva2009hyperdimensional}. 
In hypervectors, the information is distributed across multiple components rather than localized in specific dimensions as in traditional low-dimension representations. This distributed and redundant representation is often referred to as \textit{holographic representation}. This analogy comes from optical holography. A hologram is a 3D image of an object with depth, where small fragments contain information about the whole picture. Similarly, parts of an HD vector contain information about the entire encoded data. 

Two key operations in HDC are \textit{bundling} and \textit{binding}. Bundling aggregates a set of HD vectors into a single one, and {binding} encodes relationships between several pieces of information or features. 
Bundling is typically implemented as element-wise addition or majority voting.
Because of the distributed (holographic) nature of hypervectors, the resulting bundled vector retains information about each of the contributing vectors and can therefore be interpreted as an average representation of them. This property naturally supports classification tasks. HD vectors corresponding to the same class can be bundled into a so-called \textit{prototype} -- an HD vector representing this class 
-- which enables efficient similarity-based retrieval and matching.

Binding, in contrast, represents associations between elements. It is commonly implemented using element-wise multiplications or exclusive OR (XOR) between the bound hypervectors. 
For example, binding can be used to associate vectors denoting a specific property, e.g., ``pixel 1'', and its value, ``1'', so that the resulting HD representation represents ``the value of pixel 1 is~1''. The decisions and information retrieval in HDC are based on measuring similarity between hypervectors, e.g., cosine similarity that measures an angle between the vectors or Hamming distance as a more computationally efficient alternative.

HD representations offer several advantages, mostly from a hardware perspective. First, HD representations are noise-tolerant and resilient to small changes or errors in single entries. In contrast to traditional computing, HDC systems can continue functioning even if parts of the data are corrupted or incomplete. Second, all the operations on HD vectors are lightweight, i.e., they can be reduced to element-wise operations. Binary HD representations often rely on bit-shifts and XOR operations, which enable low-cost hardware implementation.

\subsection{Fundamental Procedures in HDC Systems}
Although there exists a plethora of HDC frameworks, they essentially differ in the implementation of the following fundamental procedures:

\textbf{Data transformation.} Every $d$-dimensional input vector $\set{x}\in\mathbb{R}^d$ is projected into an HD space using the selected HD transform $\theta(\set{x})$. The conventional HDC implementations rely on mapping discrete inputs into randomized binary, bipolar, or real-valued vectors using predefined lookup tables -- codebooks.
We refer the reader to \cite{Kleyko2022HyperdimensionalComputing} and \cite{kleyko2023survey_ii} for a systematic and comprehensive review of HDC implementations.

\textbf{Model initialization.} After transforming the training data, the data corresponding to each of $K$ classes are bundled, e.g., using element-wise summation, into distinct HD vectors ${\set{P}=\{\set{p}_1,\ldots,\set{p}_K\}}$ called prototypes. The prototype $\set{p}_j$ of the class $j$ can be thought of as an ``average image'' of the data corresponding to this class. That is, the prototypes are more similar to the data of the represented class than to the data of the other class.

\textbf{Inference.} 
The class of the query $\set{x}_*$ can be determined by computing the distance $\Delta_j = \mathrm{dist}(\theta(\set{x}_*),\set{p}_j)$ between its HD representation $\theta(\set{x}_*)$ and each of the prototypes $\set{p}_j,\ j={1,\ldots,K},$ initialized during the training phase. After computing the distances $\boldsymbol{\Delta}=\{\Delta_1,\ldots,\Delta_K\}$, the HDC classifier predicts the class of the query by selecting the one corresponding to the prototype with the smallest distance to the query.

\textbf{Model retraining.} The predictive performance of HDC can be further improved by calibrating the prototypes, also known as retraining. The motivation behind this procedure is that several intra-class training points can be close in the HD space, thus making the corresponding prototypes ``fuzzy''. To better separate the prototypes in the HD space, one can iterate over the training data several times by checking if any points are misclassified. HD vectors corresponding to misclassified training points are subtracted from the prototype of the incorrectly predicted class and added to the prototype of the correct one. 

\textbf{Remark:} The selection of the distance function is often determined by the type of employed HD representations. HDC frameworks with binary-valued HD vectors conventionally adopt Hamming distance due to its computational simplicity. In contrast, the frameworks with real- and integer-valued representations rely on cosine distance or dot-product. The original reason behind the adoption of these distance functions is analytically interpretable and explainable inference, as demonstrated in \cite{kanerva2009hyperdimensional}. Importantly, the choice of a distance function is ultimately determined by the HDC framework and is not limited to these conventional options.

The outlined HDC procedures provide a solid foundation for efficient FL in large-scale IIoT systems. Below, we consider the challenges of IIoT federated solutions in more detail.

\section{Challenges in Federated Learning for IIoT}

An FL-aided IIoT solution should address the unique challenges inherent to industrial environments. 
These challenges can be broadly divided into four groups: (i) resource limitations of devices and infrastructure, including computational power, energy, and cost constraints, (ii) scalability in large-scale deployments, 
and (iii) security and privacy issues. 
Below, we discuss these challenges in detail and illustrate the potential of HDC-based learning methods to resolve them.

\subsection{Resource Constraints}

Unlike more advanced edge devices that rely on powerful hardware for fast and/or real-time training and inference, IIoT devices are often limited in computational power, energy supply, and cost. Although ML procedures can be accelerated with graphical processing units (GPUs) or AI chips, such components are usually embedded into relatively expensive IIoT platforms designed for pattern recognition tasks. In addition, utilizing GPUs for ML processing comes with an increased energy consumption. Moreover, the higher cost of GPU-enabled hardware platforms can be especially critical for large-scale industrial IIoT systems, as it may extend the payback period from deploying the IIoT system into production. 

The main reason behind the dominance of GPUs in the ML domain is their ability to efficiently parallelize matrix-to-matrix multiplication, a fundamental and most resource-consuming operation in NN architectures. 
In contrast to NN-based methods, HDC is 
not 
a matrix-to-matrix multiplication-centric method and can operate \textit{completely free} of such operations. The use of bitwise operations reduces power consumption and offers significant advantages for resource-constrained IIoT devices operating on limited batteries. 
This feature also makes HDC particularly suitable for implementing on low-cost hardware such as field programmable gate arrays (FPGAs). Compared to GPU platforms, FPGAs offer a more affordable and energy-efficient alternative to deploying FL solutions. In addition, HDC requires lower memory usage and storage overhead. 
The combination of these features makes it particularly attractive for resource-constrained industrial systems. 

\subsection{Scalability in Massive Deployments}

The communication bottleneck remains a fundamental limitation of large-scale FL applications, despite being addressed by many existing FL frameworks. This problem becomes especially intense in IIoT networks, where the trade-off between massive connectivity and limited throughput leads to severe contention for the wireless resource. With tens of thousands of connected IIoT devices, the resulting congestion can directly undermine the performance of deployed FL applications. The situation is further exacerbated by the increasing density of deployments, combined with the stringent requirements on data rates and end-to-end latencies. While B5G (Beyond 5G)/6G network architectures promise to efficiently support massive IIoT deployments, their real effectiveness remains unproven yet. Meanwhile, many existing FL frameworks were designed for data-center environments and may not be directly deployable in large-scale and resource-constrained IIoT \cite{vahabi2025federated}. 

To enhance the scalability of intelligent IIoT applications, existing FL frameworks proposed communication-efficient methods of federated ML model training, such as federated dropout~\cite{niu2025fedspu} or model compression. Despite the exceptionally high performance of the NN architectures, their communication efficiency may be limited by overparameterization. In contrast, HDC prioritizes simplicity and communication efficiency and thus promises to become a more communication-efficient alternative. Due to the intrinsic redundancy of HD representations, HDC can be efficiently trained with lower precision, which may reduce the total number of transmitted bits, even given a large vector dimensionality. As a result, HDC can decrease the corresponding overhead and help mitigate the communication bottleneck.

\begin{figure*}[t!]
    \centering
    \includegraphics[width=0.7\linewidth]{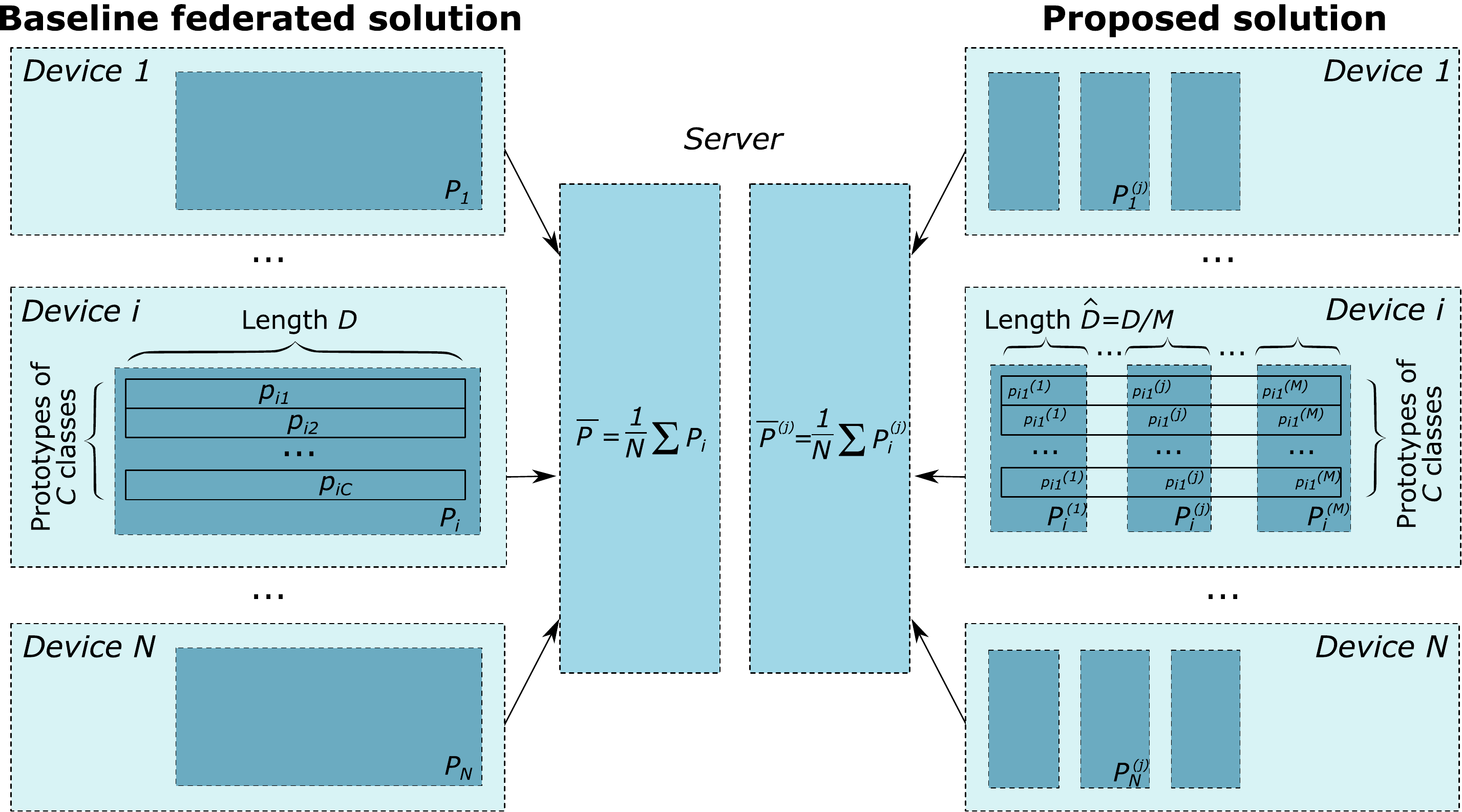}
    \caption{Architecture of proposed resource-efficient federated HDC framework for resource-constrained devices. Edge devices encode local data into hypervectors and update class prototypes. Instead of transmitting full model parameters, devices send compact prototype vectors to a central server for aggregation.}
    \label{fig:framework_arch}
\end{figure*}


\newsub{The above challenges highlight the limitations of traditional FL solutions for IIoT. With lightweight aggregation via simple algebraic updates and continuous retraining, HDC offers a more efficient solution, exceptionally suitable for ultra-low-power or latency-critical applications. IIoT challenges related to privacy and security can also be incorporated into the proposed framework, although their detailed discussion is beyond the scope of this article.
Below, we introduce our federated HDC framework, which allows optimizing both computational and communication efficiency in a federated setup.}

\section{Federated HDC Framework}

\newsub{To address the identified challenges of FL in IIoT systems, 
we propose a federated HDC framework, which offers a scalable and energy-efficient solution for inherently resource-constrained IIoT systems. Below, we discuss the integration of HDC into an FL setup to facilitate collaborative training across IIoT devices.
}

\begin{algorithm}
\SetAlgoLined
\SetAlgoNlRelativeSize{-1} 
\SetNlSty{textbf}{}{:} 
\LinesNumbered 

\KwIn{$N$ local datasets $\{\set{X}_i,\set{y}_i\}$, HDC model size $D$, HDC sub-model size $\widehat{D}$, global epochs $G$, local epochs $L$}
\KwOut{Trained local HDC models $\set{P}_1,\ldots,\set{P}_N$}

    \For{participant $i\gets 1$ \KwTo $N$}{
    Apply HD transform $\theta(\cdot)$ to local data $\set{X}_i$;\\
    \For{class $j$ $\mathbf{in}$ ${1,\ldots,K}$}{
        Find indices $\mathcal{C}^{(i)}_j$ of $i$-th participants data corresponding to class $j$;\\
        $\set{p}_j^{(i)}\gets\frac{\alpha}{|\mathcal{C}_j|}\sum_{k\in\mathcal{C}_j}\theta(\set{x_k})$; \tcp{Initialize prototype of class $j$ for participant $i$}
    }
    $\set{P}_i\gets\{\set{p}_1^{(i)},\ldots,\set{p}_K^{(i)}\}$; \tcp{Combine prototypes into HDC model}
    }
    \For{global epoch $g\gets 1$ \KwTo $G$}{
        Generate $\widehat{D}$ random indices $\set{I}\in[0,\ldots,D-1]$;\\
        \For{participant $i\gets 1$ \KwTo $N$}{
            $\tilde{\set{P}}_i$ = \texttt{LocalUpdate}($\set{P}_i,\set{I},\{\tilde{\set{X}}_i,\set{y}_i\}$);
        }
        $\tilde{\set{P}} = \frac{1}{N}\sum_{i=1}^N\tilde{\set{P}}_i$; \tcp{Perform HDC sub-model aggregation}
        \For{participant $i\gets 1$ \KwTo $N$}{
            $\set{P}_i[\set{I}]\gets\tilde{\set{P}}$; \tcp{Update retrained HDC model positions}
        }
    }
    \Return $\set{P}_1,\ldots,\set{P}_N$ \tcp{Trained HDC models of participants}
\caption{Federated HDC Training}
\label{alg:fed_training}
\end{algorithm}

\begin{algorithm}
\SetAlgoLined
\SetAlgoNlRelativeSize{-1} 
\SetNlSty{textbf}{}{:} 
\LinesNumbered 

\KwIn{Local HDC model $\set{P}$, set of indices $\set{I}$, local dataset $\{\set{X}, \set{y}\}$}
\KwOut{Updated local HDC sub-model $\tilde{\set{P}}$}

\SetKwFunction{LocalUpdate}{LocalUpdate}
\SetKwProg{Fn}{Function}{:}{}
\Fn{\LocalUpdate{$\set{P}$,$\set{I}$,$\{\set{X},\set{y}\}$}}{
    $\tilde{\set{P}}\gets\set{P}[\set{I}]$; \tcp{Form HDC sub-model}
    \For{$\ell \gets 1$ \KwTo $L$}{
        \For{data point $\{\set{x},y\}$ $\mathbf{in}$ dataset $\{\set{X},\set{y}\}$}{
            Compute distance $\boldsymbol{\Delta}=\mathrm{dist}\left(\tilde{\set{P}},\set{x}\right)$ between data point $\set{x}$ and every prototype in $\set{P}$;\\
            
            $\hat{y}=\arg\min \boldsymbol{\Delta}$;  \tcp{Select class with smallest distance}

            \If{$\hat{y}\neq y$}{
            $\set{p}_y=\set{p}_y+\alpha\cdot\Delta_y\cdot\set{x}$; \tcp{Reinforce correct classification}
            $\set{p}_{\hat{y}} = \set{p}_{\hat{y}} - \alpha\cdot\Delta_{\hat{y}}\cdot\set{x}$; \tcp{Penalize incorrect prediction}
            }
        }
    }
    \Return $\tilde{\set{P}}$ \tcp{Updated HDC sub-model}
}
\caption{Local HDC Model Update}
\label{alg:local_update}
\end{algorithm}

The envisioned framework adapts federated HDC for on-device learning with several enhancements to facilitate resource efficiency of federated training and inference on IIoT devices (see Fig.~\ref{fig:framework_arch}). We assume that the system has $N$ participants, where each participant $i$ stores a set of collected measurement vectors $\set{X}_i$ and their corresponding labels $\set{y}_i$. 
In our framework, the data vectors $\set{X}_i$ can be a set of images from the integrated camera or readings from the embedded microelectromechanical sensors (orientation, acceleration, pressure, etc.). The labels $\set{y}_i$ will depend on the IIoT application and can correspond to the current state of the IIoT device or the class of collected data.
In the following, we discuss the key procedures of our framework and how they address the resource constraints of IIoT systems. Our framework is outlined in pseudo-code in Algorithm \ref{alg:fed_training}.
Key procedures of federated HDC are the following. 

\vspace{0.5em}
\textbf{Local model initialization.} Each participant $i$ initializes a local $D$-dimensional HDC model $\set{P}_i$ from the locally stored data as discussed in Section \ref{sec:hdc_principles}. The participants must have identical HD transforms $\theta(\cdot)$ to project the original data into a common HD space. For this purpose, the participants can share a common random seed, from which they generate randomized parameters of the employed HD transform. Empirically, selecting larger dimensionality $D$ of the HDC model results in higher predictive performance but higher inference and retraining costs. Our framework can considerably reduce these while providing comparable or even higher inference performance. The initialization procedure is described in the lines 1-4 of Algorithm \ref{alg:fed_training}.

\vspace{0.5em}
\textbf{Randomized HDC sub-model retraining.} The participants randomly select a subset of $\widehat{D}$ positions of the local $D$-dimensional HDC model. This subset forms a $\widehat{D}$-dimensional HDC sub-model that is subsequently retrained during $L$ local epochs; the rest of positions are ``frozen'' and not updated. With this approach, we reduce the computational costs of retraining compared to full, $D$-dimensional HDC models as done in conventional federated HDC frameworks. In addition, this procedure may reduce overfitting as its mechanism is similar to the dropout method employed for NN training. We observe that selecting $\widehat{D}$ as a factor of $D$, that is, having $M=D/\widehat{D}$ HDC sub-models with non-overlapping positions provides the best results. A pseudo-code description of the discussed procedure is provided in Algorithm \ref{alg:local_update}.

\vspace{0.5em}
\textbf{HDC sub-model aggregation.} After $L$ local retraining epochs, the participants send their local HDC sub-models for aggregation to the parameter server. We employ the conventional FedAvg framework to combine the received users' updates into a global model (line 10 of Algorithm \ref{alg:fed_training}). The participants receive the global model and update their current HDC sub-model (lines 11-13 of Algorithm \ref{alg:fed_training}). After $G$ aggregation rounds, the federated training of the HDC model is complete. This procedure can reduce the total communication costs of the federated retraining by up to $M$ times compared to the conventional federated HDC methods.

\vspace{0.5em}
Unlike gradient-based federated learning, where communication cost scales with the number of model parameters, federated HDC exchanges fixed-length hypervectors. Consequently, communication overhead grows with the number of classes.
\begin{figure*}
\centering
    \begin{subfigure}[b]{0.45\textwidth}
        \centering
        \includegraphics[width=\textwidth]{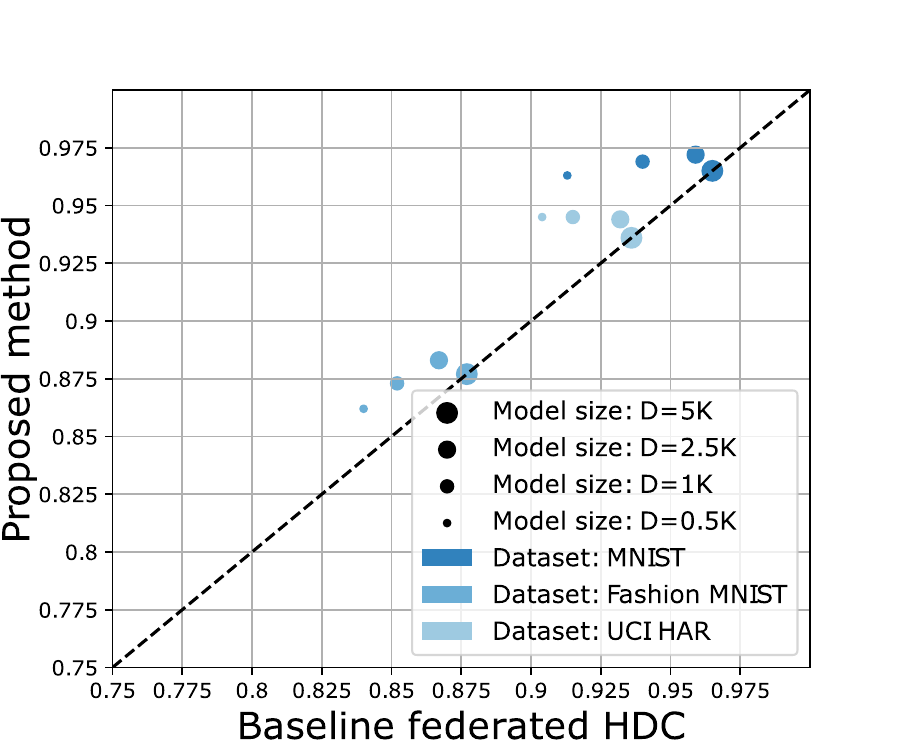}
        \caption{Performance for i.i.d. scenario.}
        \label{fig:iid}
    \end{subfigure}
    \begin{subfigure}[b]{0.45\textwidth}
        \centering
        \includegraphics[width=\textwidth]{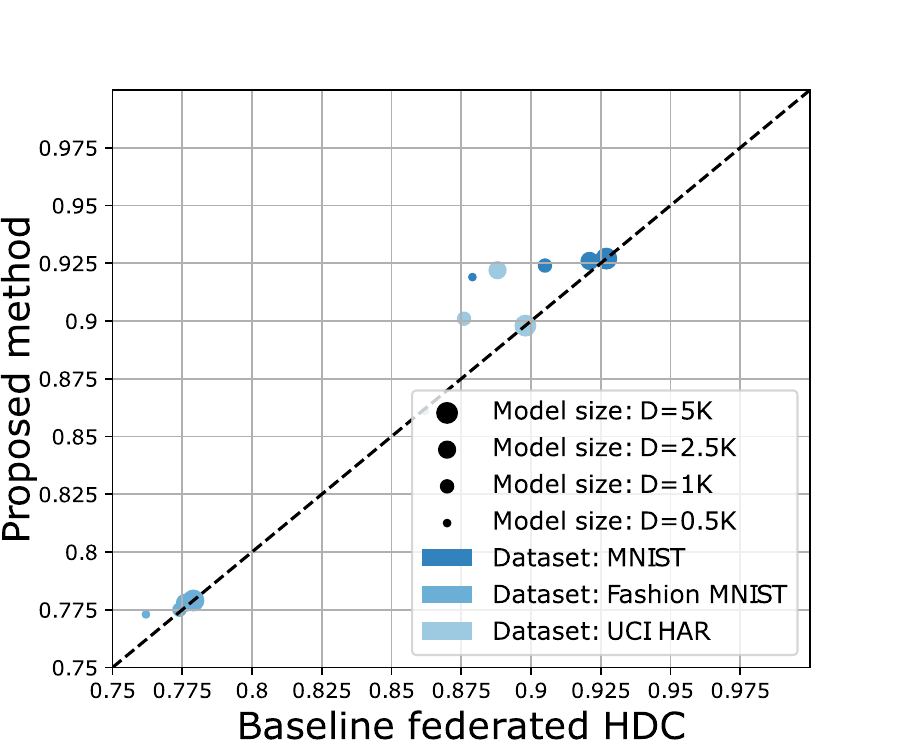}
        \caption{Performance for non-i.i.d. scenario.}
        \label{fig:non-iid}
    \end{subfigure}
    \caption{Illustrative comparison of maximum accuracy achieved by baseline and proposed federated HDC methods.}
    \label{fig:numerical_results}
\end{figure*}

The adopted federated HDC framework relies on the conventional FedAvg algorithm with a modified retraining procedure from \cite{hernandez2021onlinehd}. 
The components of this framework, however, can be enhanced according to the specifics of the applications. 
For non-i.i.d. inter-user data distributions, FedAvg may be a less favorable option than the algorithms designed to tackle heterogeneity. We argue that it is possible to adapt such algorithms as FedProx or Scaffold \cite{karimireddy2020scaffold} to the specifics of HDC models, which can improve the federated training performance. Similarly, the training convergence can be readily enhanced by employing more advanced retraining procedures, such as in \cite{zeulin2026large}.

\section{Numerical Example}

This section presents a case study of adapting our federated HDC training method for pattern recognition in image and time series data. In applied industrial problems, federated HDC can also be applied to object detection in visual inspection systems or anomaly detection in sensor networks. Here, as an example, we demonstrate the work of the envisioned systems on several community-adopted image and time-series classification datasets.

\subsection{Datasets and Data Preprocessing}

We consider a scenario with $N=20$ resource-constrained IIoT devices connected to a wireless network. We demonstrate the performance of our method on three datasets, two of which are for image classification (MNIST and Fashion MNIST) and one for time series classification (UCI HAR). For each of the considered datasets, we perform the following preprocessing steps. First, all the data points are individually normalized to a unit $\ell_2$-norm. Then, the random Fourier feature mapping (RFFM) transform is applied to every normalized data point as described in \cite{rahimi2007random}. For our HDC model, we employ OnlineHD mapping $\theta(\set{x})=\cos(\set{x}\set{W}+\set{\varphi})\cdot\sin(\set{x}\set{W})$ introduced in~\cite{hernandez2021onlinehd} to the data points after the RFFM transform. Here, $\set{W}\sim\mathcal{N}(\set{0},\set{I})$ is a $d\times D$-dimensional random projection matrix, and $\boldsymbol{\phi}\sim\mathrm{Uni}[0,2\pi]$ is a $D$-dimensional random vector. 

In this numerical demonstration, we compare two federated HDC training methods: (i) the baseline federated HDC training method and (ii) the proposed method with a dropout-inspired enhancement. For each method, we set equal learning rate $\alpha=0.01$ and the number of global epochs $G$ and local epochs $L$ to $100$ and $5$, respectively. In the baseline method, we vary the size of HDC model $D$ between $[5\mathrm{K},2.5\mathrm{K},1\mathrm{K},0.5\mathrm{K}]$, where lower-dimensional models trade smaller computational complexity and communication costs for typically lower predictive performance. In contrast, in the proposed federated HDC method, we fix the size of the HDC model to $D=5\mathrm{K}$ but update only a smaller number of randomly selected parameters $\widehat{D}=[2.5\mathrm{K},1\mathrm{K},0.5\mathrm{K}]$ at each training iteration. Therefore, the compared methods have identical computational and communication costs per federated training iteration, which makes their comparison fair.

To make our numerical comparison more extensive, we consider two scenarios with identical and non-identical class distributions between the user devices (i.i.d. and non-i.i.d., respectively). In the non-i.i.d. scenario, each device has the data of only two randomly selected classes out of ten for MNIST datasets and out of six for UCI HAR. To mitigate overfitting to local class distributions, we reduce the number of local epochs $L$ to $3$. This non-i.i.d. scenario models an extreme case, where the devices' knowledge is limited to a narrow range of data, for example, some unique proprietary data of the industrial company.

\begin{figure}
    \centering
    \includegraphics[width=\linewidth]{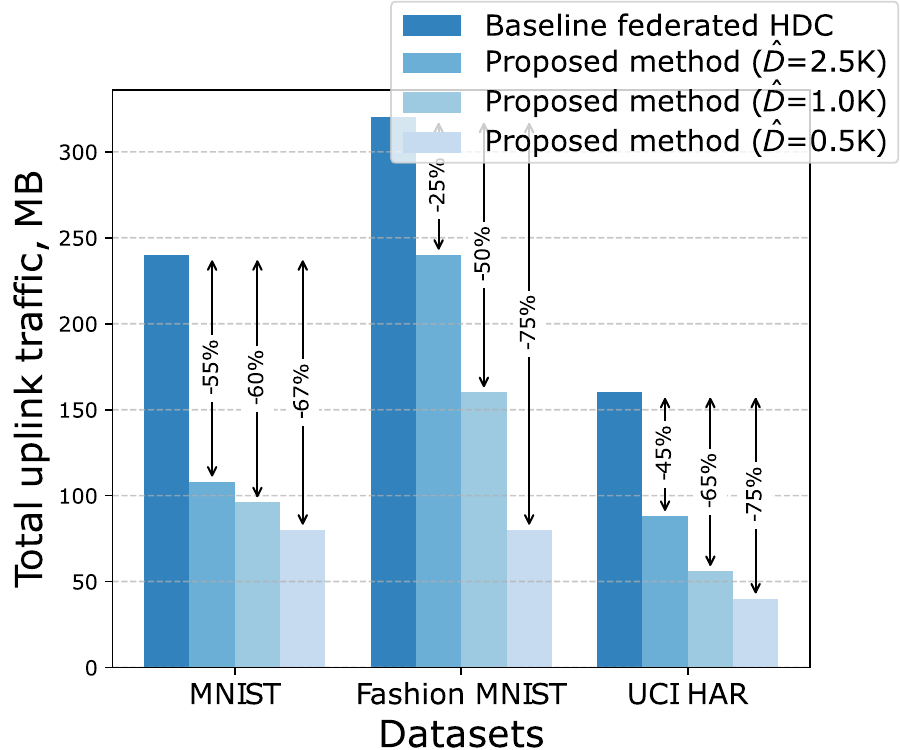}
    \caption{Total uplink traffic consumption to achieve baseline federated HDC accuracy.}
    \label{fig:uplink_traffic}
\end{figure}

\subsection{Discussion of Numerical Results}
We compare the considered federated training methods from two perspectives: classification accuracy and total uplink traffic consumption. In Fig. \ref{fig:numerical_results}, we compare the highest classification accuracy achieved by the proposed and the baseline federated HDC training methods under i.i.d. and non-i.i.d. scenarios. The dotted line corresponds to the identical performance of the considered methods. The locations above the dotted line indicate that the proposed method outperforms the baseline. The results in Fig. \ref{fig:iid} and \ref{fig:non-iid} illustrate that the proposed federated HDC method achieves even higher maximum accuracy compared to the baseline in both i.i.d. and non-i.i.d. scenarios.

The diagram in Fig. \ref{fig:uplink_traffic} compares the volumes of data transmitted by the proposed method before reaching the accuracy of the baseline for the i.i.d. scenario. These results reveal an interesting trend that selecting a smaller number of parameters $\widehat{D}$ leads to greater communication cost reduction up to $-75\%$ compared to the baseline. The demonstrated reduction is also proportional to the computational cost savings.

\begin{figure}
    \centering
    \includegraphics[width=\linewidth]{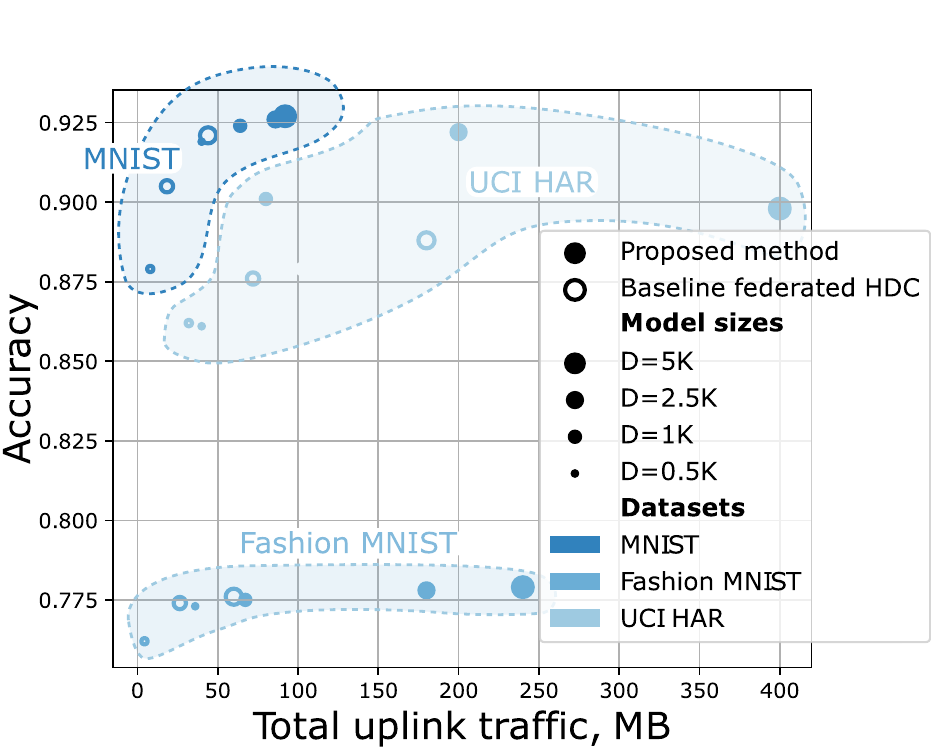}
    \caption{Comparison of maximum accuracy and volume of uplink traffic between proposed method and baseline for non-i.i.d. scenario.}
    \label{fig:accuracy_traffic_noniid}
\end{figure}

In Fig. \ref{fig:accuracy_traffic_noniid}, we provide a comparison of the maximum classification accuracy and the corresponding uplink traffic volume for the non-i.i.d. scenario. In contrast to the i.i.d. scenario, our method demonstrates higher or comparable predictive performance at the cost of higher network resource utilization. According to our results, this predominantly depends on the selected dataset. For example, for the UCI HAR dataset, the proposed method with $\widehat{D}=2.5\mathrm{K}$ considerably outperforms the baseline with $D=5\mathrm{K}$ while demonstrating traffic reduction of up to $50\%$. For MNIST and Fashion MNIST datasets, the proposed method does not outperform the baseline in predictive performance, and the uplink traffic reduction is considerably lower. Therefore, future work should explore data properties that contribute to better performance of the proposed method in non-i.i.d. scenarios and adapt it accordingly.

The proposed architecture is particularly suitable for IIoT deployments involving microcontroller-based sensing. Prototype vectors are compact and updates are infrequent; therefore, devices can participate in collaborative learning even under intermittent connectivity and limited uplink bandwidth. This makes federated HDC attractive for applications such as predictive maintenance and large-scale monitoring systems.

\section{Discussion and Conclusions}

{IIoT deployments often involve thousands of microcontroller-class sensing devices connected through bandwidth-limited wireless networks. Learning algorithms in these environments should therefore prioritize computational simplicity, low memory usage, and communication efficiency. In this article, we demonstrated that HDC -- a novel low-complexity ML framework -- can address the resource constraints of distributed IIoT systems. HDC algorithms can provide competitive predictive performance while significantly reducing computational complexity and communication overhead.
While HDC does not outperform all existing ML methods in raw accuracy, it uniquely satisfies a set of strict IIoT requirements, such as extreme resource efficiency and robustness. 
Unlike conventional techniques, HDC offers a more hardware-oriented design without matrix multiplications, which makes it especially appealing for resource-constrained deployments.}

\newsub{
HDC fills a unique niche by combining ultra-lightweight operations with robust representations and is naturally suited for incremental FL and continuous learning in resource-limited scenarios. 
The redundancy of HD representations makes HDC resilient not only to hardware-level errors such as bit flips but also to noise and incomplete data. 
These features make HDC promising for low-power, low-cost industrial environments where conventional models may fall short.
}

Moving forward, HDC components can be easily tailored to different types of data and problems, from anomaly detection in time series to image classification. The predictive performance can be further enhanced by adopting pre-trained feature extractors for data preprocessing. Furthermore, in highly resource-constrained systems, one may employ binary HD representations with inexpensive bitwise retraining and inference operations. Reducing data type resolution can additionally decrease the communications costs of the federated training. 

\bibliographystyle{ieeetr}
\bibliography{references.bib}
	
\end{document}